\pdfoutput=1

\documentclass[11pt]{article}

\usepackage{acl}

\usepackage{times}
\usepackage{latexsym}
\usepackage{amsmath}

\usepackage[T1]{fontenc}

\usepackage[utf8]{inputenc}

\usepackage{microtype}
\usepackage{multirow}
\usepackage{algorithm}
\usepackage{algorithmic}
\usepackage[switch]{lineno}
\usepackage{booktabs}
\usepackage{graphicx}
\usepackage{amssymb}

\title{Visual-Language Navigation Pretraining via Prompt-based Environmental Self-exploration}

\author{Xiwen Liang\textsuperscript{\rm 1}, Fengda Zhu\textsuperscript{\rm 2}, Lingling Li\textsuperscript{\rm 3}, Hang Xu\textsuperscript{\rm 4}, Xiaodan Liang\textsuperscript{\rm 1}\footnotemark[2] \\
  \textsuperscript{\rm 1}Shenzhen Campus of Sun Yat-sen University, Shenzhen
  \textsuperscript{\rm 2}Monash University \\
  \textsuperscript{\rm 3}Sun Yat-sen University
  \textsuperscript{\rm 4}Huawei Noah's Ark Lab}

\begin{document}
\maketitle

\renewcommand{\thefootnote}{\fnsymbol{footnote}}
\footnotetext[2]{Corresponding author.}

\begin{abstract}
Vision-language navigation (VLN) is a challenging task due to its large searching space in the environment.
To address this problem, previous works have proposed some methods of fine-tuning a large model that pretrained on large-scale datasets.
However, the conventional fine-tuning methods require extra human-labeled navigation data and lack self-exploration capabilities in environments, which hinders their generalization of unseen scenes. 
To improve the ability of fast cross-domain adaptation, we propose \textbf{Pro}mpt-\textbf{b}ased \textbf{E}nvironmental \textbf{S}elf-exploration (ProbES), which can self-explore the environments by sampling trajectories and automatically generates structured instructions via a large-scale cross-modal pretrained model (CLIP).
Our method fully utilizes the knowledge learned from CLIP to build an in-domain dataset by self-exploration without human labeling.
Unlike the conventional approach of fine-tuning, we introduce prompt-based learning to achieve fast adaptation for language embeddings, which substantially improves the learning efficiency by leveraging prior knowledge.
By automatically synthesizing trajectory-instruction pairs in any environment without human supervision and efficient prompt-based learning, our model can adapt to diverse vision-language navigation tasks, including VLN and REVERIE.
Both qualitative and quantitative results show that our ProbES significantly improves the generalization ability of the navigation model\footnote{Code will be released at \url{https://github.com/liangcici/Probes-VLN}.}.
\end{abstract}

\section{Introduction}
Teaching a robot to navigate following a natural language instruction has a broad impact in the field of human-robotic interaction.
Many related tasks have been proposed to delve into this problem.
The vision-language navigation (VLN) task~\cite{anderson2018vision} is proposed where an agent is required to navigate in a photo-realistic environment step-by-step following a natural language instruction.
Recent tasks~\cite{qi2020reverie,zhu2021soon} focus on target objects localization that asks an agent to identify an object in an unseen room.

Solving these tasks requires an agent to obtain a vision-text alignment ability that locates related objects and executes corrective actions according to the instruction.
However, collecting a large-scale VLN dataset is difficult and laborious since annotating the semantic of a trajectory within a sentence costs times of labor than annotating an image. 
Existing navigation datasets are relatively small-scale, and learning on such datasets hinders the agent to obtain a good generalization ability.
To solve this problem, EnvDrop~\cite{tan2019learning} uses a speaker model to generate instructions for sampled trajectories in unseen environments, but the generalization ability is not strong with limited vision-language understanding ability.
Recently, VLN-BERT~\cite{majumdar2020improving} introduces a visio-linguistic model pretrained on Conceptual Captions~\cite{sharma2018conceptual} dataset to learn from image-caption pairs, which are quite different from trajectory-instruction pairs from VLN.
To address this, Airbert~\cite{guhur2021airbert} constructs a large-scale in-domain pretraining dataset with image-caption pairs collected from online marketplaces such as Airbnb to finetune ViLBERT.
However, Airbert collects image captioning data on websites, which are still far from the scenario of vision-language navigation.
Different from previous methods that collect human-labeled data to train a navigation model, we suggest that automatically generating instruction-trajectory pairs by self-exploration for pretraining not only helps the model obtain better generalization ability but also achieves fast adaptation to downstream tasks.

\begin{figure*}[!t]
    \begin{center}
        \includegraphics[width=0.99\linewidth]{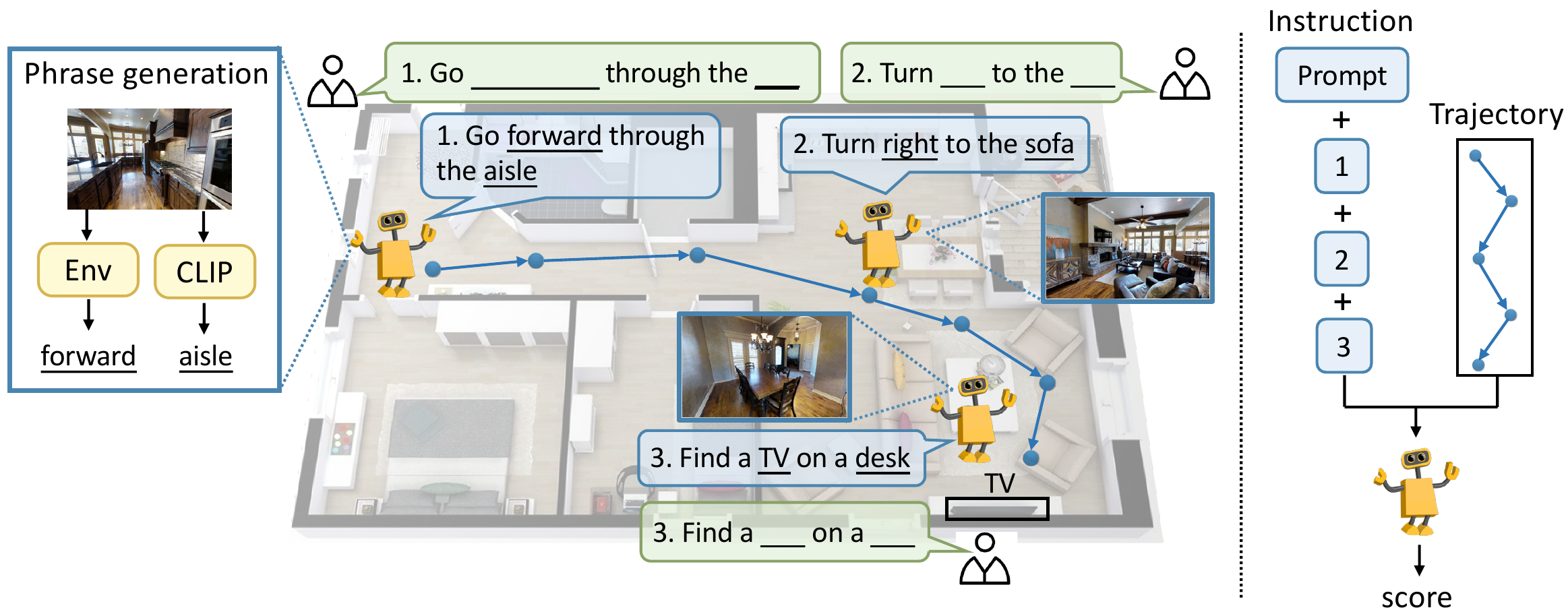}
    \end{center}
\vspace{-0.6em}
\caption{A demonstration of our prompt-based environmental self-exploration. In the left blue box, we sample trajectories from the environment and generate candidate phrases by a pretrained CLIP model. Then we fill templates by movements and the generated phrases during self-exploration. At last, we use the generated instruction-trajectory samples for pretraining.}
\label{fig:overview}
\vspace{-4mm}
\end{figure*}

In this paper, we propose a method named prompt-based environmental self-exploration (ProbES) that generates navigation data with prior knowledge automatically and adapts pretrained model quickly to VLN tasks. 
An overview of our proposed framework is shown in Figure~\ref{fig:overview}.
By using this method, a pretrained visio-linguistic model is able to adapt to the VLN task automatically and efficiently.
Specifically, we build an in-domain dataset by self-exploration without labeling or crawler.
To build such a dataset. we first generate templates by masking visual and action words in labeled instructions.
Then, we sample trajectories in the training environment.
A pretrained CLIP~\cite{radford2021learning} model is used to recognize rooms and objects in the sampled trajectories and match described phrases with them.
We construct instructions by filling the matched phrases into sampled templates.
By leveraging the prior knowledge learned by CLIP, we are able to build a dataset automatically with rich semantic information.
Meanwhile, finetuning the whole pretrained model is time-consuming, we adopt prompt tuning~\cite{li2021prefix,liu2021gpt,liu2021ptuning}, a lightweight alternative to finetuning. 
Our prompt-based method can distill task-relevant knowledge from pretrained model and achieve fast adaption to downstream tasks.
We evaluate ProbES on R2R~\cite{anderson2018vision} and REVERIE~\cite{qi2020reverie} datasets by discriminative and generative settings.
Results show that ProbES can match or surpass the performance of finetuning with substantially less training time.

To sum up, our main contributions are as follows:
(1) We propose ProbES, a novel self-exploration method to automatically build an in-domain dataset that reduces the domain gap between the pretraining dataset and VLN tasks without human labeling; (2) Compared with finetuning large pretrained model, our proposed prompt tuning can achieve fast adaptation; (3) Experiments are conducted on R2R and REVERIE datasets with generative and discriminative settings, and results indicate that our proposed ProbES can achieve better or comparable performance. Besides, our generated data can be used as augmented data which improves the generalization ability of the model.

\section{Related Work}

\noindent\textbf{Vision-and-Language Navigation.}
Anderson et al.~\cite{anderson2018vision} proposed the first Vision-Language Navigation (VLN) benchmark combining real imagery~\cite{Matterport3D} and natural language navigation instructions. To solve this task, Wang et al.~\cite{wang2020soft} proposed a novel SERL model to learn reward functions from the expert distribution.
And combining imitation learning and reinforcement learning~\cite{wang2019reinforced} has been proved to be beneficial for VLN.
Since the VLN dataset is relatively small-scale, some works propose augmentation approaches~\cite{fried2018speaker, tan2019learning, liu2021vision} to improve robustness.
Auxiliary losses~\cite{majumdar2020improving,zhu2020vision,liang2021contrastive} is used to take advantage of the additional training signals derived from the semantic information.
Some pretraining methods~\cite{huang2019transferable, hao2020towards} have been proposed to learn generic cross-modal representations. This is further extended to a recurrent model that significantly improves sequential action prediction~\cite{hong2021vln}.
However, the limited number of environments in pretraining constrain the generalization ability to unseen scenarios.
Most related to this work, VLN-BERT~\cite{majumdar2020improving} transfers knowledge from abundant, but out-of-domain image-text data to improve path-instruction matching. In contrast, we not only propose an effective method to build an in-domain dataset by sampling trajectory and generating instructions with templates, but also present a prompt-based pretraining strategy to improve VLN.

\noindent\textbf{Vision-and-Language Pretraining.} Vision-and-language pretraining has made great progress in recent years. Inspired by BERT~\cite{devlin2018bert}, much work has extended it to process visual tokens and pretrain on large-scale image-text pairs for learning generic visio-linguistic representations. Previous research introduces one-stream BERT models and two-stream BERT models. The former directly perform inter-modal grounding~\cite{li2019visualbert,su2019vl,alberti2019fusion,li2020unicoder,chen2020uniter,zhou2020unified,li2020oscar}, while two-stream models process both visual and textual inputs in separate streams, and then fuse the two modalities in a later stage~\cite{lu2019vilbert,tan2019lxmert}. These models are often pretrained with self-supervised objectives akin to those in BERT: masked language modeling, masked object classification, and sentence-image alignment. In this work, the architecture of the ProbES model is structural similar to ViLBERT~\cite{lu2019vilbert}. We make several VLN-specific adaptations to ViLBERT so that pretrained weights can be transferred to initialize large portions of the model. Different from VLN-BERT which fine-tunes a ViLBERT on instruction-trajectory pairs to measure their compatibility in beam search setting, we introduce prompt tuning, which only tunes the continuous prompts.

\begin{figure*}[!t]
    \begin{center}
        \includegraphics[width=0.99\linewidth]{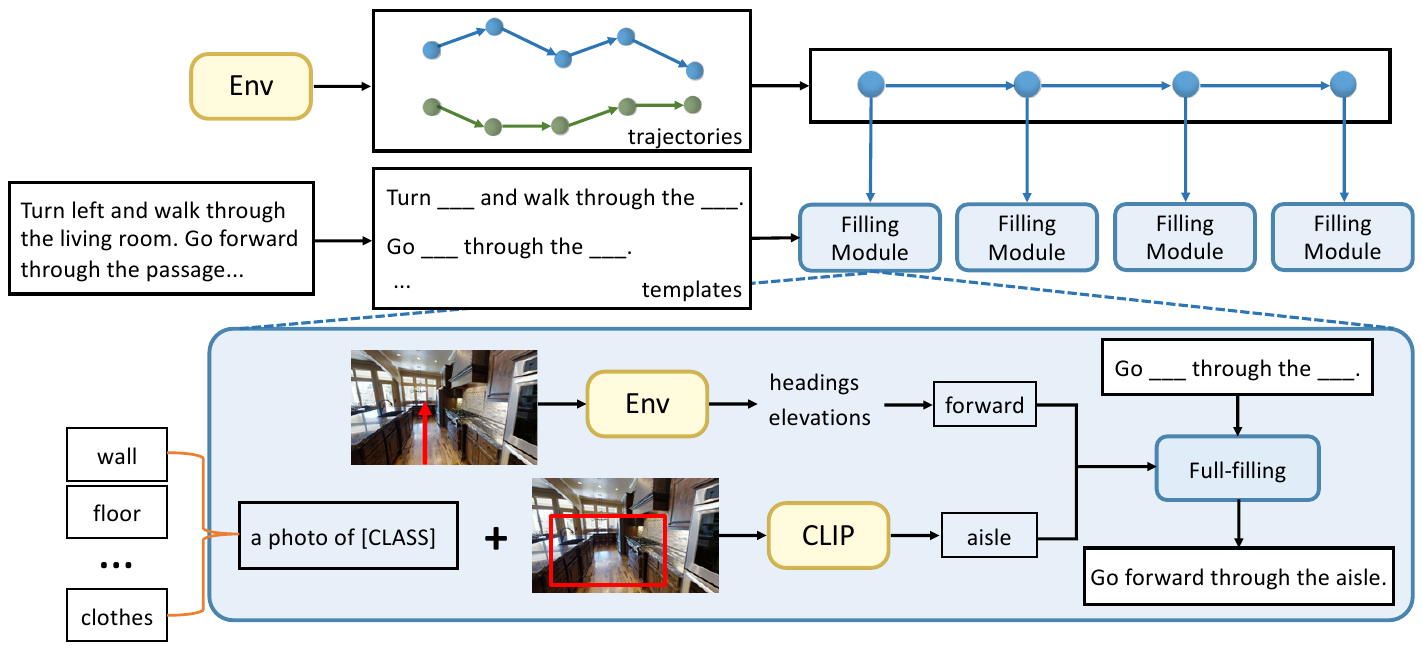}
    \end{center}
\vspace{-0.6em}
\caption{A detailed demonstration of the prompt-based full-filling process. We first sample trajectories from the environment, and generate templates by masking objects and actions. For each step of a trajectory, we generate candidate tokens for objects by CLIP and actions by the environment. Then we full-fill the template with candidate tokens by the rules as introduced in Sec.~\ref{sec:method_gen_instr}}
\label{fig:framework}
\vspace{-4mm}
\end{figure*}

\noindent\textbf{Prompting.}
Natural language prompting freezes pretrained models and reformats the natural language input with example prompts. GPT-3~\cite{Brown2020Language} introduces in-context learning, using manually designed and discrete text prompts. Sun et al.~\cite{sun2020conditioned} also leverage prompts as keywords to control the sentiment or topic of the generated sentence. AutoPrompt~\cite{shin2020autoprompt} searches for a sequence of discrete trigger words and concatenates it with each input to elicit sentiment or factual knowledge from a masked LM.
Different from the discrete text prompt, some methods examine continuous prompts (a.k.a. soft prompts) that perform prompting directly in the embedding space of the model. Prefix-Tuning~\cite{li2021prefix} prepends a sequence of continuous task-specific vectors as virtual tokens to the input. \cite{zhong2021factual,qin2021learning, hambardzumyan2021warp} introduce continuous templates following manual prompt templates. P-tuning~\cite{liu2021gpt} uses continuous prompts which are learned by inserting trainable variables into the embedded input. Ptr~\cite{han2021ptr} adopts manually crafted sub-templates and generates complete templates by logic rules. In ProbES, we prepend continuous task-specific vectors to the embedding of the input instruction and directly tune the embeddings of these vectors. After prompt tuning, the model can be adapted to VLN and REVERIE tasks.

\section{Prompt-based Environmental Self-Exploration (ProbES)} 

\subsection{Vision-Language Navigation}
The Vision-and-Language Navigation (VLN) task gives a global natural sentence $I=\{w_0 ,..., w_l\}$ as an instruction, where $w_i$ is a word token while the $l$ is the length of the sentence. The instruction consists of step-by-step guidance toward the goal. At step $t$, the agent observes a panoramic view $O_{t}=\{o_{t,i}\}_{i=1}^{36}$ as the vision input, which is composed of 36 RGB image views. Each of these views consists of image feature $v_i$ and an orientation description ($sin\ \theta_{t,i}$, $cos\ \theta_{t,i}$, $sin\ \phi_{t,i}$, $cos\ \phi_{t,i}$). Candidates in the panoramic action space consist of $k$ neighbours of the current node in the navigation graph and a stop action.

\subsection{Instruction Generation with Templates} \label{sec:method_gen_instr}

We first generate templates from instructions in the R2R dataset. Then we sample trajectories in the training environment.
We generate the candidate noun phrases and actionable verbs for the sampled trajectories and full-fill the templates by the above words.
A detailed demonstration of our instruction generation module is shown in Fig.~\ref{fig:framework}.

\noindent\textbf{Generating Templates}
We collect phrases and replace these phrases in human-annotated navigation instruction with blank masks to generate templates.
Different from the Airbert~\cite{guhur2021airbert} that only extracts noun phrases, we also mask action words like `left', `right', 'forward', and `around'.
We denote the $O_{mask}$ as the mask for an object and $A_{mask}$ is the mask for an action.
The generated templates are like `Turn $A_{mask}$ and walk past $O_{mask}$. Once out, walk $A_{mask}$ $O_{mask}$. Stop once you reach $O_{mask}$'.
More examples are shown in Table~\ref{tab:sample}.

\noindent\textbf{Sampling Trajectories and Actions}
We first sample the trajectories in the Matterport~\cite{Matterport3D} Environment.
We randomly sample the starting and ending positions, and collect tracks with lengths of less than 8 hops. 
Then we obtain the corresponding actions of each trajectory by first-person movement.
If the agent chooses the front navigable position to move, we generate a `forward' action.
If the agent chooses the back navigable position to move, we generate an `around' action.
Otherwise, if the agent selects the right front navigable position to move for the next step, we generate an action sequence like \{`right', `forward'\}, which is used to fill actionable verbs during instruction generation.

\noindent\textbf{Full-filling Template with Prior Knowledge} 
Prior knowledge is the key to generating high-quality data without human labeling.
ProbES introduces CLIP, a powerful vision-language alignment model learned from a large-scale image-caption dataset.
To generate structured augmentation data, we full-fill the templates with phrases that describe the sampled trajectory and actions.
A trajectory is denoted as $\{v_1, v_2,...,v_n\}$, where $v_i$ represents an observation viewpoint.
We introduce CLIP~\cite{radford2021learning} to select candidate phrases $c$ and match them to each view $v_i$.
We first embed the sentence `a photo of [$c_{noun}$]' by CLIP, where the $c_{noun}$ represents the noun-phrase candidates (room or object classes labeled in Matterport dataset).
Then we embed the view image by the vision encoder of CLIP and calculate the similarity of the language embedding and vision embedding.
We select the candidate with the highest matching score for the view $v_i$.
Each view has two matched candidates, one for the detected room and another for an object. Then the description $c_i$ of this view is written in 3 formats randomly: `[room]', `[object]' or `[room] with [object]'.
Since trajectories are sampled in the environment, we can obtain actionable verbs $a_i$ between two viewpoints via comparing headings and elevations.

\begin{table*}[]
    \centering
    \caption{Examples of generated templates.}
    \vspace{-2mm}
    \resizebox{0.98\linewidth}{!}{
    {\renewcommand{\arraystretch}{1}
    \begin{tabular}{l|l}
        \toprule
         & Templates\\
        \midrule
        1 & Walk $A_{mask}$ $O_{mask}$ and stop on $O_{mask}$.  \\
        2 & Head $A_{mask}$ until you pass $O_{mask}$ with $O_{mask}$ the turn $A_{mask}$ and wait by $O_{mask}$.\\
        3 & Walk past $O_{mask}$ and to $O_{mask}$. Walk in $O_{mask}$ and stop. \\
        4 & Turn $A_{mask}$ and walk through $O_{mask}$. Exit $O_{mask}$, turn $A_{mask}$ and walk $A_{mask}$ $O_{mask}$. Stop in $O_{mask}$. \\
        5 & Go $A_{mask}$ $O_{mask}$, and go $A_{mask}$. Take $A_{mask}$ into $O_{mask}$. Stop behind $O_{mask}$. \\
        6 & Leave $O_{mask}$ and go through $O_{mask}$. Walk towards $O_{mask}$ to $O_{mask}$. Stand in $O_{mask}$.\\
        \bottomrule
    \end{tabular}}
    }
    \vspace{-2mm}
    \label{tab:sample}
\end{table*}

We randomly select a template with the same or a close number of $O_{mask}$ as the number of viewpoints in the sampled trajectory.
The template has a sequence of object masks $\{O_{mask,1}, O_{mask,2},...,O_{mask,i}\}$ and a sequence of action masks $\{A_{mask,1}, A_{mask,2}, ..., A_{mask,j}\}$.
Lengths of object masks and action masks are denoted as $l$ and $n$ respectively. The number of object masks and action masks is roughly balanced.
Let $n_v$ be the number of viewpoints in a sampled trajectory. Then the generated captions of this trajectory is written as $\{c_1, c_2, ..., c_{n_v}\}$.
We full-fill the templates by the following rules:
1) if $n_v \geq l$, we randomly sample $l$ captions and fill the $O_{mask}$ in the template sequentially; 
2) if $n_v \textless l$, we randomly sample the $O_{mask}$ and use all the caption phrases to fill them.
After filling phrases, we can identify which viewpoint $A_{mask,i}$ may appear since viewpoints of $O_{mask,j}$ near it are already known. For example, if the template is like `$O_{mask,1} A_{mask,1} O_{mask,2}$' and captions of $v_1$ and $v_2$ are used to fill $O_{mask,1}$ and $O_{mask,2}$ respectively, then $A_{mask,1}$ is the sampled action between $v_1$ and $v_2$. In this way, we use generated actionable verbs to full-fill the templates and get final instructions.
By the above method, we can generate diverse instructions without human labeling.

\subsection{Prompt-based Architecture}
Prompt tuning has been found effective on many natural language understanding (NLU) tasks.
Motivated by this, we introduce a prompt-based architecture to achieve fast adaptation on the self-exploration dataset (e.g., Conceptual Captions) and downstream tasks.
The architecture is ViLBERT-like and equipped with a prompt encoder for prompt tuning.

Given an instruction-trajectory pair, the visual and textual features can be extracted by the visual encoder $E_v$ and textual encoder $E_x$ in ViLBERT respectively.
Especially, the textual input has two parts: prompt sequence $\{p_1, ..., p_n\}$  and word sequence $\{x_1, ..., x_m\}$, where $p$ and $x$ indicate a pseudo prompt token and a word token of a generated instruction respectively. $n$ and $m$ represent lengths of the prompt sequence and word sequence respectively.

We embed prompt sequence by the prompt encoder $E_p$ and embed word sequence by the textual encoder $E_x$ as follows:
\begin{equation}
    \begin{split}
        e_{p,1},...,e_{p,n} &= E_p(p_1,..., p_n) \\
        e_{x,1},...,e_{x,m} &= E_x(x_1),..., E_x(x_m),
    \end{split}
\end{equation}
where $E_p$ is composed of a LSTM head followed by a MLP head.
Then the textual embedding is mapped to $e_t = \{e_{p,1},...,e_{p,n},e_{x,1},...,e_{x,m}\}$, where $e_{p,1},...,e_{p,n}$ are trainable embedding tensors and enable us to find better continous prompts.
Let $e_v$ be denoted as visual embedding produced by visual encoder $E_v$.
$e_t$ and $e_v$ are then passed to the co-attention transformer similar to ViLBERT.
Then in the prompt tuning process, we only train $E_p$ and fix the parameters of $E_x$ for the language stream. For the vision stream, since the trajectory is represented as a sequence of panoramic image regions, which is different from VLMs pretrained on image-caption pairs, we also update the visual embedding during prompt tuning. The visual embedding contains image embedding and location embedding.

We sample hard negative paths based on distance in the environment for an instruction-trajectory pair, and the model is trained to choose the best path among them. 

\subsection{Downstream Tasks Adaptation}
Our model can adapt to diverse downstream navigation tasks, including VLN, a step-by-step navigation task, and REVERIE, an object-oriented navigation task.
In the step-by-step navigation task, our model receives an instruction sentence and navigates following the commands in the instruction sequentially.
In the object navigation task, our model receives an object description and explores the house to find an object.

Also, our model can be adapted to both discriminative and generative navigation settings.
In the discriminative setting, our model receives both an instruction and the observation sequence to represent a navigation trajectory and then output a score.
In the generative setting, our model receives instruction and predicts actions sequentially.

\section{Experiments}

\subsection{Experimental Setup}

\begin{table*}[]
    \centering
    \caption{Comparison with previous methods in the generative setting on the R2R dataset.}
    \vspace{-2mm}
    \resizebox{0.98\linewidth}{!}{
    {\renewcommand{\arraystretch}{1}
    \begin{tabular}{l|cccc|cccc|cccc}
        \toprule
        \multirow{2}{*}{} & \multicolumn{4}{c|}{Val Seen} & \multicolumn{4}{c|}{Val Unseen} & \multicolumn{4}{c}{Test Unseen} \\
        \cline{2-13}
         & TL & NE$\downarrow$ & SR$\uparrow$ & SPL$\uparrow$ & TL & NE$\downarrow$ & SR$\uparrow$ & SPL$\uparrow$ & TL & NE$\downarrow$ & SR$\uparrow$ & SPL$\uparrow$ \\
        \midrule
        Seq2Seq-SF & 11.33 & 6.01 & 39 & - & 8.39 & 7.81 & 22 & - & 8.13 & 7.85 & 20 & 18 \\
        Speaker-Follower & - & 3.36 & 66 & - & - & 6.62 & 35 & - & 14.82 & 6.62 & 35 & 28 \\
        PRESS & 10.57 & 4.39 & 58 & 55 & 10.36 & 5.28 & 49 & 45 & 10.77 & 5.49 & 49 & 45  \\
        EnvDrop & 11.00 & 3.99 & 62 & 59 & 10.70 & 5.22 & 52 & 48 & 11.66 & 5.23 & 51 & 47 \\
        PREVALENT & 10.32 & 3.67 & 69 & 65 & 10.19 & 4.71 & 58 & 53 & 10.51 & 5.30 & 54 & 51 \\
        Rec (no init. OSCAR) & 9.78 & 3.92 & 62 & 59 & 10.31 & 5.10 & 50 & 46 & 11.15 & 5.45 & 51 & 47 \\
        Rec (OSCAR) & 10.79 & 3.11 & 71 & 67 & 11.86 & 4.29 & 59 & 53 & 12.34 & 4.59 & 57 & 53 \\
        Rec (PREVALENT) & 11.13 & \textbf{2.90} & \textbf{72} &  \textbf{68} & 12.01 & \textbf{3.93} & \textbf{63} & \textbf{57} & 12.35 & \textbf{4.09} & \textbf{63} & \textbf{57} \\
        \midrule
        Rec (ViLBERT) & 11.16 & 2.54 & 75 & 71 & 12.44 & 4.20 & 60 & 54 & - & - & - & - \\
        Rec (VLN-BERT) & 10.95 & 3.37 & 68 & 64 & 11.33 & 4.19 & 60 & 55 & - & - & - & - \\
        Rec (ProbES) & 10.75 & {\color{blue}\textbf{2.95}}  &  {\color{blue}\textbf{73}} & {\color{blue}\textbf{69}} & 11.58 & {\color{blue}\textbf{4.03}}  & {\color{blue}\textbf{61}} & {\color{blue}\textbf{55}} & 12.43 & {\color{blue}\textbf{4.20}}  & {\color{blue}\textbf{62}} & {\color{blue}\textbf{56}}  \\
        \bottomrule
    \end{tabular}}}
    \vspace{-2mm}
    \label{tab:r2r_single_run}
\end{table*}

\begin{table*}[]
    \Huge
    \centering
    \caption{Comparison with previous methods on navigation and object localization on the REVERIE dataset.}
    
    \vspace{-2mm}
    \resizebox{1.0\linewidth}{!}{
    {\renewcommand{\arraystretch}{1}
    \begin{tabular}{l|cccc|cc|cccc|cc|cccc|cc}
        \toprule
        \multirow{3}{*}{} & \multicolumn{6}{c|}{Val Seen} & \multicolumn{6}{c|}{Val Unseen} & \multicolumn{6}{c}{Test Unseen} \\
        \cline{2-19}
         & \multicolumn{4}{c|}{Navigation} & \multirow{2}{*}{RGS} & \multirow{2}{*}{RGSPL} & \multicolumn{4}{c|}{Navigation} & \multirow{2}{*}{RGS} & \multirow{2}{*}{RGSPL} & \multicolumn{4}{c|}{Navigation} & \multirow{2}{*}{RGS} & \multirow{2}{*}{RGSPL} \\
         & SR & OSR & SPL & TL & & & SR & OSR & SPL & TL & & & SR & OSR & SPL & TL & & \\
        \midrule
        Seq2Seq-SF & 29.59 & 35.70 & 24.01 &  12.88 & 18.97 & 14.96 & 4.20 & 8.07 & 2.84 & 11.07 &  2.16 & 1.63 & 3.99 & 6.88 & 3.09 & 10.89 & 2.00 & 1.58 \\
        RCM & 23.33 & 29.44 & 21.82 & 10.70 & 16.23 & 15.36 & 9.29 & 14.23 & 6.97 & 11.98 & 4.89 & 3.89 & 7.84 & 11.68 & 6.67 & 10.60 & 3.67 & 3.14 \\
        SMNA & 41.25 & 43.29 & 39.61 & 7.54 & 30.07 & 28.98 & 8.15 & 11.28 & 6.44 & 9.07 & 4.54 & 3.61 & 5.80 & 8.39 & 4.53 & 9.23 & 3.10 & 2.39 \\
        FAST-MATTN & \textbf{50.53} &  \textbf{55.17} & \textbf{45.50} & 16.35 & 31.97 & 29.66 & 14.40 & 28.20 & 7.19 & 45.28 & 7.84 & 4.67 & 19.88 & 30.63 & 11.61 & 39.05 & 11.28 & 6.08 \\
        Rec (OSCAR) & 39.85 & 41.32 & 35.86 & 12.85 & 24.46 & 22.28 & 25.53 & 27.66 & 21.06 & 14.35 & 14.20 & 12.00 & 24.62 & 26.67 & 19.48 & 14.88 &  12.65 & 10.00 \\
        \midrule
        Rec (ViLBERT) & 43.64 & 45.61 & 37.86 & 15.75 & 31.69 & 27.58 & 24.57 & 29.91 & 19.81 & 17.83 & 15.14 & 12.15 & 22.17 & 25.51 & 17.28 & 18.22 & 12.87 & 10.00 \\
        Rec (VLN-BERT) & 41.11 & 42.87 &  35.55 & 15.62 & 28.39 & 24.99 & 25.53 & 29.42 & 20.51 & 16.94 & 16.42 & 13.29 & 23.57 & 26.83 & 18.73 & 17.63 & 14.24 & 11.63 \\
        Rec (ProbES) & 46.52 & 48.49 & 42.44 & 13.59 & \textbf{33.66} & \textbf{30.86} & \textbf{27.63} & \textbf{33.23} & \textbf{22.75} & 18.00 & \textbf{16.84} & \textbf{13.94} & \textbf{24.97} & \textbf{28.23} & \textbf{20.12} & 17.43 & \textbf{15.11} & \textbf{12.32} \\
        \bottomrule
    \end{tabular}}}
    \vspace{-3mm}
    \label{tab:reverie}
\end{table*}

We experiment with our proposed ProbES on two downstream tasks: goal-oriented navigation task (R2R~\cite{anderson2018vision}), and object-oriented navigation task (REVERIE~\cite{qi2020reverie}).
ProbES can be easily applied to discriminative and generative models for these two tasks.

\noindent\textbf{Evaluation Metrics} 
A large number of metrics are used to evaluate models in VLN, such as Trajectory Length (TL), the trajectory length in meters, Navigation Error (NE), the navigation error in meters, Oracle Success Rate (OR), the rate if the agent successfully stops at the closest point, Success Rate (SR), the success rate of reaching the goal, and Success rate weighted by (normalized inverse) Path Length (SPL)~\cite{anderson2018on}. VLN task regard SR and SPL as the primary metric, and the REVERIE task regard RGS and RGSPL as the primary metric.

\noindent\textbf{Implementation Details} 
Our training process is divided into two steps: Firstly, we pretrain our model on our generated self-exploration training set with prompt tuning for only 10 epochs. After that, we adapt our model to the downstream discriminative VLN task with only ranking loss for 20 epochs. The batch size is set as 64 and the learning rate is $4\times10^{-5}$.
The generative navigation settings are the same as Recurrent VLN-BERT on both R2R and REVERIE.
During pretraining, we use ProbES to 50k instruction-trajectory pairs.
We use 32 NVIDIA V100 GPUs for pretraining and 8 GPUs for adaptation. Experiments with generative settings are conducted on a V100 GPU.

\begin{table}[]
    \centering
    \caption{Results by comparing ProbES with VLN-BERT in discriminative setting.}
    \vspace{-2mm}
    \resizebox{0.95\linewidth}{!}{
    {\renewcommand{\arraystretch}{1}
    \begin{tabular}{l|ccccc}
        \toprule
        \multirow{2}{*}{} & \multicolumn{5}{c}{Val Unseen} \\
        \cline{2-6}
         & TL & NE$\downarrow$ & OSR$\uparrow$ & SR$\uparrow$ & SPL$\uparrow$ \\
        \midrule
        VLN-BERT & 9.60 & 4.10 & 69.22 & 59.26 & 55 \\
        ProbES & 9.50 & 4.05 & 68.24 & 60.28 & 56 \\
        \bottomrule
    \end{tabular}}}
    \vspace{-4mm}
    \label{tab:r2r_val}
\end{table}

\subsection{Comparison to state-of-the-art Methods}

In this section, we compare our model with previous state-of-the-art methods.  We compare the ProbES with two baselines (ViLBERT and VLN-BERT built on Recurrent VLN-Bert) and five other methods. A brief description of previous models is as followed:
1) Seq2Seq: A sequence to sequence model reported in~\cite{anderson2018vision};
2) Speaker-Follower~\cite{fried2018speaker}: a method introduces a data augmentation approach and panoramic action space;
3) PRESS~\cite{li2019robust}: a conventional fine-tuning method with stochastic instruction sampling;
4) EnvDrop~\cite{tan2019learning}: a method augment data with environmental dropout;
5) Recurrent VLN-Bert~\cite{hong2021vln} on three different settings: OSCAR and ViLBERT pretrained on out-of-domain data, VLN-BERT pretrained on R2R.
We compare the models on three splits in the R2R dataset: validation seen house, validation unseen house, and testing (where the houses are also unseen).
We  also compare ProbES with Seq2Seq, RCM~\cite{wang2019reinforced}, SMNA~\cite{ma2019self}, FAST-MATTN~\cite{qi2020reverie}, Recurrent VLN-Bert~\cite{hong2021vln} on OSCAR on REVERIE dataset.

\noindent\textbf{Results on R2R} 
We compare ProbES with previous state-of-the-art methods on the R2R dataset in the generative setting, which predicts actions sequentially, as shown in Table~\ref{tab:r2r_single_run}.
In the validation seen split, compared to VLN-BERT under the same setting, our ProbES achieves 5\% improvement on SR and 5\% improvement on SPL.
In the validation unseen split, we achieve 1\% improvement on SR compared to VLN-BERT.
In the testing split, ProbES shows competitive results.
Note that the PREVALENT backbone is pretrained on an in-domain R2R dataset with scene features and fine-tuned with an additional action prediction task in a generative setting while ProbES does not use labeled R2R data or augmented data generated by speaker~\cite{fried2018speaker}.

\noindent\textbf{Results in Discriminative Setting} 
We compare ProbES with VLN-BERT in the discriminative setting, which outputs scores for instruction-trajectory pairs, as in Table~\ref{tab:r2r_val}.
In the validation unseen split, our method outperforms VLN-BERT, which indicates ProbES is able to improve the generalization ability for unseen scenes.

\begin{figure*}[!t]
    \begin{center}
        \includegraphics[width=0.99\linewidth]{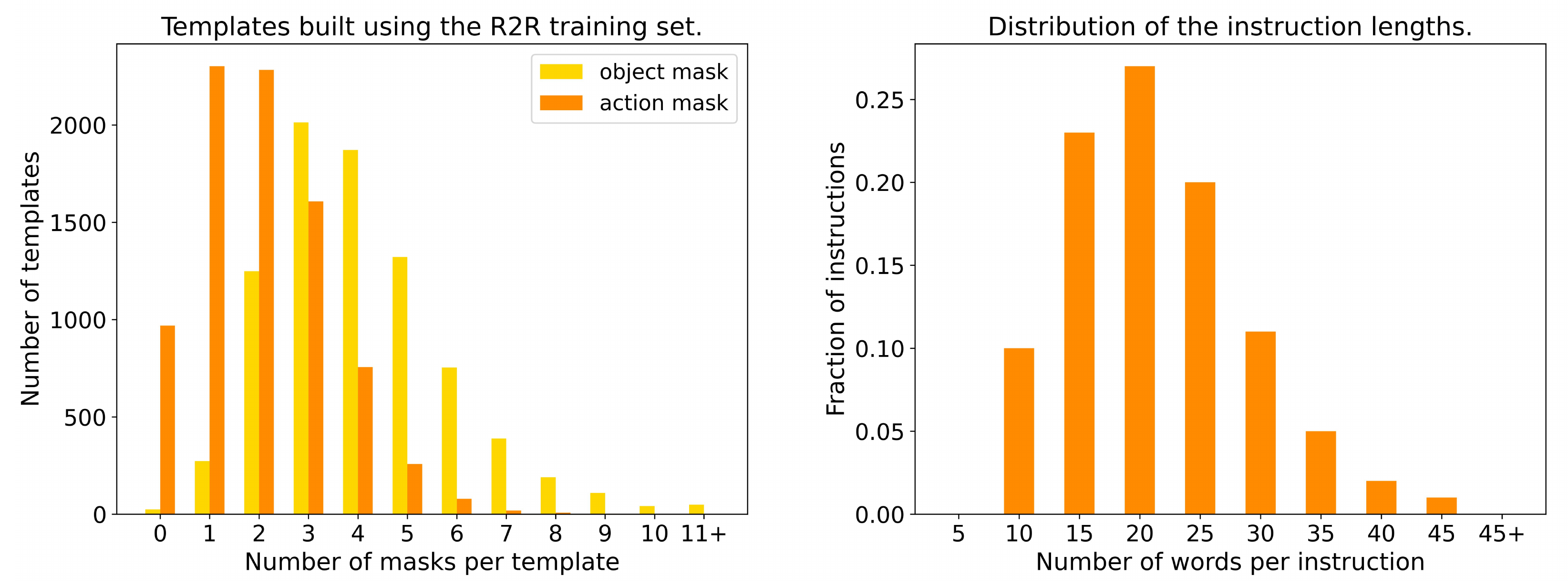} %
    \end{center}
\vspace{-3mm}
\caption{Statistical analysis of generated instructions.}
\label{fig:statistics}
\vspace{-0.8em}
\end{figure*}

\begin{figure*}[!t]
    \begin{center}
        \includegraphics[width=0.99\linewidth]{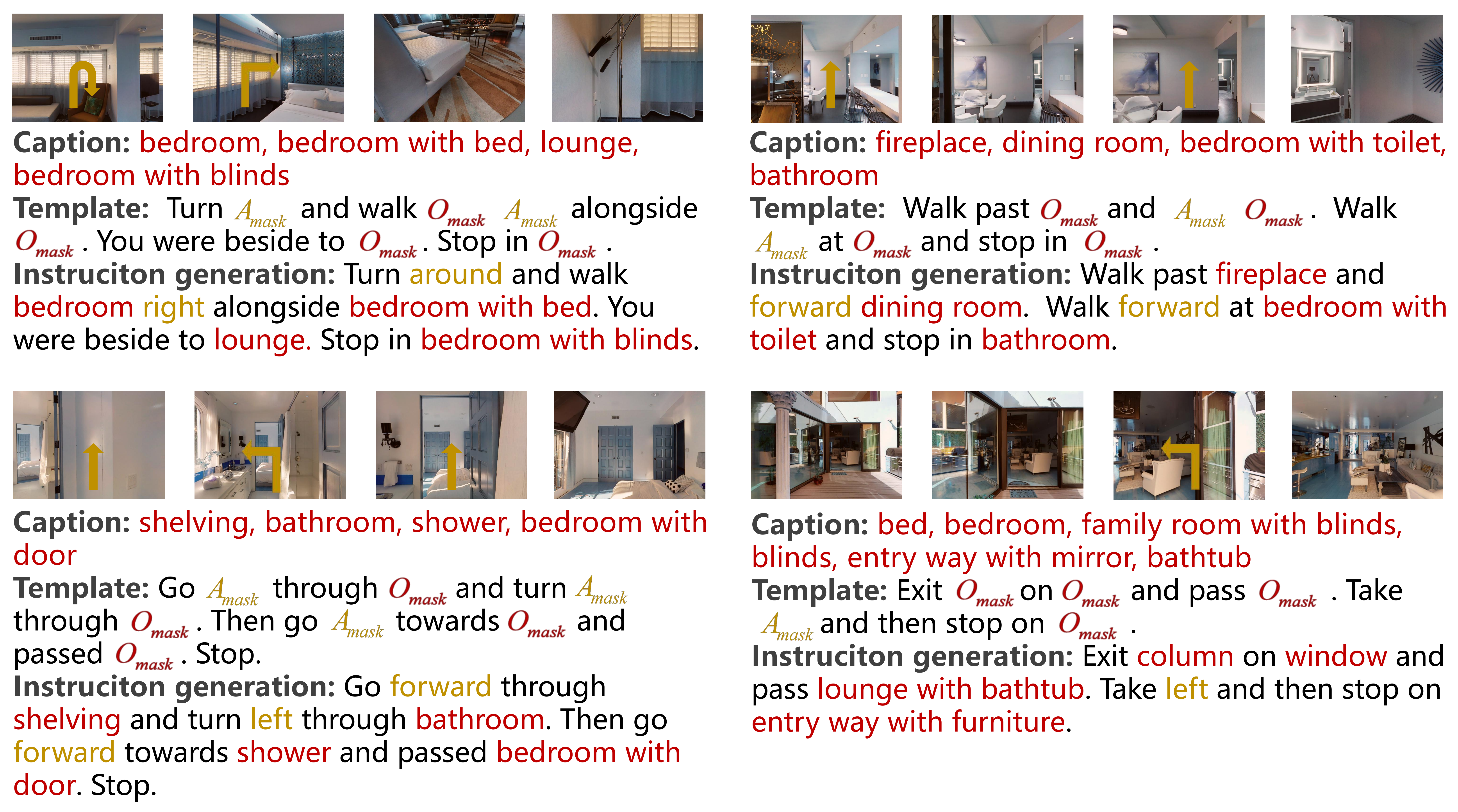} %
    \end{center}
\vspace{-15pt}
\caption{Visualization of instructions generated with templates.}
\label{fig:visualization}
\vspace{-4mm}
\end{figure*}

\noindent\textbf{Results on REVERIE} 
We compare ProbES with previous state-of-the-art methods on the REVERIE dataset, as shown in Table~\ref{tab:reverie}.
In the validation unseen split, we achieve 0.42\% improvement on RGS and 0.65\% improvement on RGSPL.
In the testing split, ProbES achieves 0.87\% improvement on RGS and 0.69\% improvement on RGSPL.
We can see that ProbES benefits from prompt tuning with our generated instruction-trajectory pairs.

\begin{table}[]
    \centering
    \caption{Ablation of different modules during pretraining and finetuning. }
    \vspace{-2mm}
    \resizebox{0.98\linewidth}{!}{
    {\renewcommand{\arraystretch}{1}
    \begin{tabular}{l|ccc|cc|cc}
        \toprule
         & \multicolumn{3}{c|}{Our data} & \multicolumn{2}{c|}{R2R} & \multicolumn{2}{c}{SR on Val} \\
         & PT & FT & Mask & Mask & Rank & Seen & Unseen \\
        \midrule
        1 & - & - & - & - & \checkmark & 55.4 & 39.5 \\
        2 & - & - & - & \checkmark & \checkmark & \textbf{70.2} & 59.3 \\
        3 & - & - & \checkmark & - & \checkmark & 69.1 & 57.9 \\
        3 & - & \checkmark & - & - & \checkmark & 68.7 & 59.0 \\
        4 & \checkmark & - & - & - & \checkmark & 68.4 & \textbf{60.3} \\
        \bottomrule
    \end{tabular}}}
    \vspace{-4mm}
    \label{tab:module}
\end{table}

\subsection{Ablation Study}

\noindent\textbf{Ablation of Learning Strategies.}
In Table~\ref{tab:module}, we ablate the performance gains from different learning strategies.
PT and FT represent prompt tuning and fine-tuning respectively.
Mask and Rank stand for masked multi-modal modeling loss and the ranking loss for path-selection task.
We regard the model finetuned by ranking loss as our baseline.

The masked multi-modal modeling loss on our data and R2R data are able to improve the performance. And finetuning on our data is able to improve generalization ability since the success rate in the validation unseen split gets 1.1\% improvement and achieves 59.0\%.
At last, we discover that pretraining on our data with prompt tuning improves the baseline performance by 20.8\% in the validation unseen split, achieving the best performance.
Our model outperforms the model fine-tuned on R2R dataset by 1.1\% in unseen split, indicating that ProbES improves the generalization ability of the navigation model.

\noindent\textbf{Ablation of Instruction Generation.} 
Table~\ref{tab:ablation-geninstr} introduces comprehensive ablation experiments showing the impact of key steps in the strategy of generating instructions, and the experiments are performed in the baseline model: IL$+$RL from EnvDrop~\cite{tan2019learning}.
Class indicates classes we use to feed into CLIP. M and P/O represent classes from Matterport and Place365/Objects365 datasets respectively. $G_{Template}$ denotes the strategy used to generate templates. `ours' denote the strategy shown in Sec~\ref{sec:method_gen_instr}. For $S_{Template}$, `random' and `match' indicate sampling a template randomly and choosing a template with the same number of masks as the number of viewpoints.

As shown in Table \ref{tab:ablation-geninstr}, randomly selecting template without considering the number of masked tokens degrades the performance and introduces more noise in the data.
Results show that equipped with our generated data (Row 3) improves the performance by a large margin.
The model of using the rooms and objects from Places365~\cite{zhou2017places} and Objects365~\cite{Objects365} (Row 4) performs worse than which uses the rooms and objects from Matterport. 
We infer from that Places365 and Objects365 contain many outdoor scenes and objects which are not suitable for VLN.

\begin{table}[]
    \centering
    \caption{Comparison of different strategies during generating instructions. }
    \vspace{-2mm}
    \resizebox{0.98\linewidth}{!}{
    {\renewcommand{\arraystretch}{1}
    \begin{tabular}{l|cc|c|cc|cc}
        \toprule
         & \multicolumn{2}{c|}{Class} & \multicolumn{1}{c|}{$G_{Template}$} & \multicolumn{2}{c|}{$S_{Instruction}$} & \multicolumn{2}{c}{SR on Val}\\
         & M & P/O & ours & random & match & Seen & Unseen \\
        \midrule
        1 & - & - & - &  - &  - &  55.3 & 46.5 \\
        2 & \checkmark & - & \checkmark & \checkmark & - & 59.8 & 49.4 \\
        3 & \checkmark & -  & \checkmark & - & \checkmark& 60.5 & 50.7 \\
        4 & - & \checkmark & \checkmark & \checkmark & - & 59.8 & 48.9 \\
        \bottomrule
    \end{tabular}}}
    \vspace{-4mm}
    \label{tab:ablation-geninstr}
\end{table}

\subsection{Qualititiva Analysis}

\noindent\textbf{Visualization of Data Distribution}
Figure~\ref{fig:statistics} presents a statistical analysis of our generated instructions.
We can see from the left figure that the number of object masks are larger than that of action masks, indicating that instructions contain more rich information generated by CLIP from sampled observations.
The right figure shows the distribution of the instruction lengths.
The lengths of most of the instructions range from 10 to 30, which matches the R2R dataset.
The easy samples and hard samples in our generated instructions are balanced.

\noindent\textbf{Visualization of Trajectory-instruction pairs}
Here we provide visualization of the data generated by ProbES.
Figure~\ref{fig:visualization} shows the instruction-trajectory samples generated with our strategy.
For each sample, we visualize observations of the trajectory, captions generated with CLIP, the selected template, and the final instruction generated by ProbES.
Generated object classes fit observed scenes well, thus we can infer that CLIP is able to extract key information from the observation.
Also, our method can select a suitable template and generate diverse instructions that describe observations of trajectories correctly.
The length of our generated instruction ranges from 1 to 3 sentences, which matches the data distribution of the R2R dataset.

\section{Conclusion}
In this work, we first introduce an effective way to generate in-domain data for pretraining the VLN model: leveraging a large pretrained CLIP model to generate captions for each viewpoint and sampling actions in the environment. Experiments show that the domain gap between pretraining data and VLN tasks can be mitigated. We also propose a prompt-based architecture, which introduces prompt tuning to adapt the pretrained model fastly. Our proposed ProbES achieves better results compared to baseline on both R2R and REVERIE datasets, and ablations show the contribution of each module and the effectiveness of the generated data.

\section*{Acknowledgement}
This work was supported in part by National Natural Science Foundation of China (NSFC) No.61976233, Guangdong Province Basic and Applied Basic Research (Regional Joint Fund-Key) Grant No.2019B1515120039, Guangdong Outstanding Youth Fund (Grant No. 2021B1515020061), Shenzhen Fundamental Research Program (Project No. RCYX20200714114642083, No. JCYJ20190807154211365) and CAAI-Huawei MindSpore Open Fund.  We thank MindSpore for the partial support of this work, which is a new deep learning computing framwork\footnote{https://www.mindspore.cn/}, and supported by Guangdong Provincial Key Laboratory of Fire Science and Intelligent Emergency Technology, Guangzhou 510006, China.

\bibliography{anthology,ref}

\begin{thebibliography}{43}
\expandafter\ifx\csname natexlab\endcsname\relax\def\natexlab#1{#1}\fi

\bibitem[{{Alberti} et~al.(2019){Alberti}, {Ling}, {Collins}, and
  {Reitter}}]{alberti2019fusion}
Chris {Alberti}, Jeffrey {Ling}, Michael {Collins}, and David {Reitter}. 2019.
\newblock Fusion of detected objects in text for visual question answering.
\newblock In \emph{EMNLP-IJCNLP}, pages 2131--2140.

\bibitem[{{Anderson} et~al.(2018){Anderson}, {Chang}, {Chaplot}, {Dosovitskiy},
  {Gupta}, {Koltun}, {Kosecka}, {Malik}, {Mottaghi}, {Savva}, and
  {Zamir}}]{anderson2018on}
Peter {Anderson}, Angel~X. {Chang}, Devendra~Singh {Chaplot}, Alexey
  {Dosovitskiy}, Saurabh {Gupta}, Vladlen {Koltun}, Jana {Kosecka}, Jitendra
  {Malik}, Roozbeh {Mottaghi}, Manolis {Savva}, and Amir~Roshan {Zamir}. 2018.
\newblock On evaluation of embodied navigation agents.
\newblock \emph{arXiv preprint arXiv:1807.06757}.

\bibitem[{Anderson et~al.(2018)Anderson, Wu, Teney, Bruce, Johnson,
  S{\"u}nderhauf, Reid, Gould, and Van Den~Hengel}]{anderson2018vision}
Peter Anderson, Qi~Wu, Damien Teney, Jake Bruce, Mark Johnson, Niko
  S{\"u}nderhauf, Ian Reid, Stephen Gould, and Anton Van Den~Hengel. 2018.
\newblock Vision-and-language navigation: Interpreting visually-grounded
  navigation instructions in real environments.
\newblock In \emph{CVPR}, pages 3674--3683.

\bibitem[{{Brown} et~al.(2020){Brown}, {Mann}, {Ryder}, {Subbiah}, {Kaplan},
  {Dhariwal}, {Neelakantan}, {Shyam}, {Sastry}, {Askell}, {Agarwal},
  {Herbert-Voss}, {Krueger}, {Henighan}, {Child}, {Ramesh}, {Ziegler}, {Wu},
  {Winter}, {Hesse}, {Chen}, {Sigler}, {Litwin}, {Gray}, {Chess}, {Clark},
  {Berner}, {McCandlish}, {Radford}, {Sutskever}, and
  {Amodei}}]{Brown2020Language}
Tom~B. {Brown}, Benjamin {Mann}, Nick {Ryder}, Melanie {Subbiah}, Jared
  {Kaplan}, Prafulla {Dhariwal}, Arvind {Neelakantan}, Pranav {Shyam}, Girish
  {Sastry}, Amanda {Askell}, Sandhini {Agarwal}, Ariel {Herbert-Voss}, Gretchen
  {Krueger}, Tom {Henighan}, Rewon {Child}, Aditya {Ramesh}, Daniel~M.
  {Ziegler}, Jeffrey {Wu}, Clemens {Winter}, Christopher {Hesse}, Mark {Chen},
  Eric {Sigler}, Mateusz {Litwin}, Scott {Gray}, Benjamin {Chess}, Jack
  {Clark}, Christopher {Berner}, Sam {McCandlish}, Alec {Radford}, Ilya
  {Sutskever}, and Dario {Amodei}. 2020.
\newblock Language models are few-shot learners.
\newblock In \emph{NeurIPS}, volume~33, pages 1877--1901.

\bibitem[{{Chang} et~al.(2017){Chang}, {Dai}, {Funkhouser}, {Halber},
  {Niebner}, {Savva}, {Song}, {Zeng}, and {Zhang}}]{Matterport3D}
Angel {Chang}, Angela {Dai}, Thomas {Funkhouser}, Maciej {Halber}, Matthias
  {Niebner}, Manolis {Savva}, Shuran {Song}, Andy {Zeng}, and Yinda {Zhang}.
  2017.
\newblock Matterport3d: Learning from rgb-d data in indoor environments.
\newblock In \emph{3DV}, pages 667--676.

\bibitem[{Chen et~al.(2020)Chen, Li, Yu, El~Kholy, Ahmed, Gan, Cheng, and
  Liu}]{chen2020uniter}
Yen-Chun Chen, Linjie Li, Licheng Yu, Ahmed El~Kholy, Faisal Ahmed, Zhe Gan,
  Yu~Cheng, and Jingjing Liu. 2020.
\newblock Uniter: Universal image-text representation learning.
\newblock In \emph{ECCV}, pages 104--120.

\bibitem[{Devlin et~al.(2018)Devlin, Chang, Lee, and
  Toutanova}]{devlin2018bert}
Jacob Devlin, Ming-Wei Chang, Kenton Lee, and Kristina Toutanova. 2018.
\newblock Bert: Pre-training of deep bidirectional transformers for language
  understanding.
\newblock \emph{arXiv preprint arXiv:1810.04805}.

\bibitem[{{Fried} et~al.(2018){Fried}, {Hu}, {Cirik}, {Rohrbach}, {Andreas},
  {Morency}, {Berg-Kirkpatrick}, {Saenko}, {Klein}, and
  {Darrell}}]{fried2018speaker}
Daniel {Fried}, Ronghang {Hu}, Volkan {Cirik}, Anna {Rohrbach}, Jacob
  {Andreas}, Louis-Philippe {Morency}, Taylor {Berg-Kirkpatrick}, Kate
  {Saenko}, Dan {Klein}, and Trevor {Darrell}. 2018.
\newblock Speaker-follower models for vision-and-language navigation.
\newblock In \emph{NeurIPS}, volume~31, pages 3314--3325.

\bibitem[{Guhur et~al.(2021)Guhur, Tapaswi, Chen, Laptev, and
  Schmid}]{guhur2021airbert}
Pierre-Louis Guhur, Makarand Tapaswi, Shizhe Chen, Ivan Laptev, and Cordelia
  Schmid. 2021.
\newblock Airbert: In-domain pretraining for vision-and-language navigation.
\newblock In \emph{ICCV}, pages 1634--1643.

\bibitem[{Hambardzumyan et~al.(2021)Hambardzumyan, Khachatrian, and
  May}]{hambardzumyan2021warp}
Karen Hambardzumyan, Hrant Khachatrian, and Jonathan May. 2021.
\newblock Warp: Word-level adversarial reprogramming.
\newblock \emph{arXiv preprint arXiv:2101.00121}.

\bibitem[{Han et~al.(2021)Han, Zhao, Ding, Liu, and Sun}]{han2021ptr}
Xu~Han, Weilin Zhao, Ning Ding, Zhiyuan Liu, and Maosong Sun. 2021.
\newblock Ptr: Prompt tuning with rules for text classification.
\newblock \emph{arXiv preprint arXiv:2105.11259}.

\bibitem[{Hao et~al.(2020)Hao, Li, Li, Carin, and Gao}]{hao2020towards}
Weituo Hao, Chunyuan Li, Xiujun Li, Lawrence Carin, and Jianfeng Gao. 2020.
\newblock Towards learning a generic agent for vision-and-language navigation
  via pre-training.
\newblock In \emph{CVPR}, pages 13137--13146.

\bibitem[{Hong et~al.(2021)Hong, Wu, Qi, Rodriguez-Opazo, and
  Gould}]{hong2021vln}
Yicong Hong, Qi~Wu, Yuankai Qi, Cristian Rodriguez-Opazo, and Stephen Gould.
  2021.
\newblock Vln bert: A recurrent vision-and-language bert for navigation.
\newblock In \emph{CVPR}, pages 1643--1653.

\bibitem[{Huang et~al.(2019)Huang, Jain, Mehta, Ku, Magalhaes, Baldridge, and
  Ie}]{huang2019transferable}
Haoshuo Huang, Vihan Jain, Harsh Mehta, Alexander Ku, Gabriel Magalhaes, Jason
  Baldridge, and Eugene Ie. 2019.
\newblock Transferable representation learning in vision-and-language
  navigation.
\newblock In \emph{ICCV}, pages 7404--7413.

\bibitem[{Li et~al.(2020{\natexlab{a}})Li, Duan, Fang, Gong, and
  Jiang}]{li2020unicoder}
Gen Li, Nan Duan, Yuejian Fang, Ming Gong, and Daxin Jiang. 2020{\natexlab{a}}.
\newblock Unicoder-vl: A universal encoder for vision and language by
  cross-modal pre-training.
\newblock In \emph{AAAI}, volume~34, pages 11336--11344.

\bibitem[{Li et~al.(2019)Li, Yatskar, Yin, Hsieh, and Chang}]{li2019visualbert}
Liunian~Harold Li, Mark Yatskar, Da~Yin, Cho-Jui Hsieh, and Kai-Wei Chang.
  2019.
\newblock Visualbert: A simple and performant baseline for vision and language.
\newblock \emph{arXiv preprint arXiv:1908.03557}.

\bibitem[{Li and Liang(2021)}]{li2021prefix}
Xiang~Lisa Li and Percy Liang. 2021.
\newblock Prefix-tuning: Optimizing continuous prompts for generation.
\newblock \emph{arXiv preprint arXiv:2101.00190}.

\bibitem[{{Li} et~al.(2019){Li}, {Li}, {Xia}, {Bisk}, {Celikyilmaz}, {Gao},
  {Smith}, and {Choi}}]{li2019robust}
Xiujun {Li}, Chunyuan {Li}, Qiaolin {Xia}, Yonatan {Bisk}, Asli {Celikyilmaz},
  Jianfeng {Gao}, Noah~A. {Smith}, and Yejin {Choi}. 2019.
\newblock Robust navigation with language pretraining and stochastic sampling.
\newblock In \emph{EMNLP-IJCNLP}, pages 1494--1499.

\bibitem[{Li et~al.(2020{\natexlab{b}})Li, Yin, Li, Zhang, Hu, Zhang, Wang, Hu,
  Dong, Wei et~al.}]{li2020oscar}
Xiujun Li, Xi~Yin, Chunyuan Li, Pengchuan Zhang, Xiaowei Hu, Lei Zhang, Lijuan
  Wang, Houdong Hu, Li~Dong, Furu Wei, et~al. 2020{\natexlab{b}}.
\newblock Oscar: Object-semantics aligned pre-training for vision-language
  tasks.
\newblock In \emph{ECCV}, pages 121--137.

\bibitem[{Liang et~al.(2021)Liang, Zhu, Zhu, Lin, Wang, and
  Liang}]{liang2021contrastive}
Xiwen Liang, Fengda Zhu, Yi~Zhu, Bingqian Lin, Bing Wang, and Xiaodan Liang.
  2021.
\newblock Contrastive instruction-trajectory learning for vision-language
  navigation.
\newblock \emph{arXiv preprint arXiv:2112.04138}.

\bibitem[{Liu et~al.(2021{\natexlab{a}})Liu, Zhu, Chang, Liang, Ge, and
  Shen}]{liu2021vision}
Chong Liu, Fengda Zhu, Xiaojun Chang, Xiaodan Liang, Zongyuan Ge, and Yi-Dong
  Shen. 2021{\natexlab{a}}.
\newblock Vision-language navigation with random environmental mixup.
\newblock In \emph{ICCV}, pages 1644--1654.

\bibitem[{Liu et~al.(2021{\natexlab{b}})Liu, Ji, Fu, Du, Yang, and
  Tang}]{liu2021ptuning}
Xiao Liu, Kaixuan Ji, Yicheng Fu, Zhengxiao Du, Zhilin Yang, and Jie Tang.
  2021{\natexlab{b}}.
\newblock P-tuning v2: Prompt tuning can be comparable to fine-tuning
  universally across scales and tasks.
\newblock \emph{arXiv preprint arXiv:2110.07602}.

\bibitem[{Liu et~al.(2021{\natexlab{c}})Liu, Zheng, Du, Ding, Qian, Yang, and
  Tang}]{liu2021gpt}
Xiao Liu, Yanan Zheng, Zhengxiao Du, Ming Ding, Yujie Qian, Zhilin Yang, and
  Jie Tang. 2021{\natexlab{c}}.
\newblock Gpt understands, too.
\newblock \emph{arXiv preprint arXiv:2103.10385}.

\bibitem[{Lu et~al.(2019)Lu, Batra, Parikh, and Lee}]{lu2019vilbert}
Jiasen Lu, Dhruv Batra, Devi Parikh, and Stefan Lee. 2019.
\newblock Vilbert: Pretraining task-agnostic visiolinguistic representations
  for vision-and-language tasks.
\newblock \emph{arXiv preprint arXiv:1908.02265}.

\bibitem[{Ma et~al.(2019)Ma, Lu, Wu, AlRegib, Kira, Socher, and
  Xiong}]{ma2019self}
Chih-Yao Ma, Jiasen Lu, Zuxuan Wu, Ghassan AlRegib, Zsolt Kira, Richard Socher,
  and Caiming Xiong. 2019.
\newblock Self-monitoring navigation agent via auxiliary progress estimation.
\newblock \emph{arXiv preprint arXiv:1901.03035}.

\bibitem[{{Majumdar} et~al.(2020){Majumdar}, {Shrivastava}, {Lee}, {Anderson},
  {Parikh}, and {Batra}}]{majumdar2020improving}
Arjun {Majumdar}, Ayush {Shrivastava}, Stefan {Lee}, Peter {Anderson}, Devi
  {Parikh}, and Dhruv {Batra}. 2020.
\newblock Improving vision-and-language navigation with image-text pairs from
  the web.
\newblock In \emph{ECCV}, pages 259--274.

\bibitem[{Qi et~al.(2020)Qi, Wu, Anderson, Wang, Wang, Shen, and
  Hengel}]{qi2020reverie}
Yuankai Qi, Qi~Wu, Peter Anderson, Xin Wang, William~Yang Wang, Chunhua Shen,
  and Anton van~den Hengel. 2020.
\newblock Reverie: Remote embodied visual referring expression in real indoor
  environments.
\newblock In \emph{CVPR}, pages 9982--9991.

\bibitem[{Qin and Eisner(2021)}]{qin2021learning}
Guanghui Qin and Jason Eisner. 2021.
\newblock Learning how to ask: Querying lms with mixtures of soft prompts.
\newblock \emph{arXiv preprint arXiv:2104.06599}.

\bibitem[{{Radford} et~al.(2021){Radford}, {Kim}, {Hallacy}, {Ramesh}, {Goh},
  {Agarwal}, {Sastry}, {Askell}, {Mishkin}, {Clark}, {Krueger}, and
  {Sutskever}}]{radford2021learning}
Alec {Radford}, Jong~Wook {Kim}, Chris {Hallacy}, Aditya {Ramesh}, Gabriel
  {Goh}, Sandhini {Agarwal}, Girish {Sastry}, Amanda {Askell}, Pamela
  {Mishkin}, Jack {Clark}, Gretchen {Krueger}, and Ilya {Sutskever}. 2021.
\newblock Learning transferable visual models from natural language
  supervision.
\newblock In \emph{ICML}, pages 8748--8763.

\bibitem[{Shao et~al.(2019)Shao, Li, Zhang, Peng, Yu, Zhang, Li, and
  Sun}]{Objects365}
Shuai Shao, Zeming Li, Tianyuan Zhang, Chao Peng, Gang Yu, Xiangyu Zhang, Jing
  Li, and Jian Sun. 2019.
\newblock Objects365: A large-scale, high-quality dataset for object detection.
\newblock In \emph{ICCV}, pages 8429--8438.

\bibitem[{Sharma et~al.(2018)Sharma, Ding, Goodman, and
  Soricut}]{sharma2018conceptual}
Piyush Sharma, Nan Ding, Sebastian Goodman, and Radu Soricut. 2018.
\newblock Conceptual captions: A cleaned, hypernymed, image alt-text dataset
  for automatic image captioning.
\newblock In \emph{ACL}, pages 2556--2565.

\bibitem[{Shin et~al.(2020)Shin, Razeghi, Logan~IV, Wallace, and
  Singh}]{shin2020autoprompt}
Taylor Shin, Yasaman Razeghi, Robert~L Logan~IV, Eric Wallace, and Sameer
  Singh. 2020.
\newblock Autoprompt: Eliciting knowledge from language models with
  automatically generated prompts.
\newblock \emph{arXiv preprint arXiv:2010.15980}.

\bibitem[{Su et~al.(2019)Su, Zhu, Cao, Li, Lu, Wei, and Dai}]{su2019vl}
Weijie Su, Xizhou Zhu, Yue Cao, Bin Li, Lewei Lu, Furu Wei, and Jifeng Dai.
  2019.
\newblock Vl-bert: Pre-training of generic visual-linguistic representations.
\newblock \emph{arXiv preprint arXiv:1908.08530}.

\bibitem[{Sun and Lai(2020)}]{sun2020conditioned}
Fan-Keng Sun and Cheng-I Lai. 2020.
\newblock Conditioned natural language generation using only unconditioned
  language model: An exploration.
\newblock \emph{arXiv preprint arXiv:2011.07347}.

\bibitem[{Tan and Bansal(2019)}]{tan2019lxmert}
Hao Tan and Mohit Bansal. 2019.
\newblock Lxmert: Learning cross-modality encoder representations from
  transformers.
\newblock \emph{arXiv preprint arXiv:1908.07490}.

\bibitem[{Tan et~al.(2019)Tan, Yu, and Bansal}]{tan2019learning}
Hao Tan, Licheng Yu, and Mohit Bansal. 2019.
\newblock Learning to navigate unseen environments: Back translation with
  environmental dropout.
\newblock \emph{arXiv preprint arXiv:1904.04195}.

\bibitem[{Wang et~al.(2020)Wang, Wu, and Shen}]{wang2020soft}
Hu~Wang, Qi~Wu, and Chunhua Shen. 2020.
\newblock Soft expert reward learning for vision-and-language navigation.
\newblock In \emph{ECCV}, pages 126--141.

\bibitem[{Wang et~al.(2019)Wang, Huang, Celikyilmaz, Gao, Shen, Wang, Wang, and
  Zhang}]{wang2019reinforced}
Xin Wang, Qiuyuan Huang, Asli Celikyilmaz, Jianfeng Gao, Dinghan Shen,
  Yuan-Fang Wang, William~Yang Wang, and Lei Zhang. 2019.
\newblock Reinforced cross-modal matching and self-supervised imitation
  learning for vision-language navigation.
\newblock In \emph{CVPR}, pages 6629--6638.

\bibitem[{Zhong et~al.(2021)Zhong, Friedman, and Chen}]{zhong2021factual}
Zexuan Zhong, Dan Friedman, and Danqi Chen. 2021.
\newblock Factual probing is [mask]: Learning vs. learning to recall.
\newblock \emph{arXiv preprint arXiv:2104.05240}.

\bibitem[{Zhou et~al.(2017)Zhou, Lapedriza, Khosla, Oliva, and
  Torralba}]{zhou2017places}
Bolei Zhou, Agata Lapedriza, Aditya Khosla, Aude Oliva, and Antonio Torralba.
  2017.
\newblock Places: A 10 million image database for scene recognition.
\newblock \emph{PAMI}.

\bibitem[{Zhou et~al.(2020)Zhou, Palangi, Zhang, Hu, Corso, and
  Gao}]{zhou2020unified}
Luowei Zhou, Hamid Palangi, Lei Zhang, Houdong Hu, Jason Corso, and Jianfeng
  Gao. 2020.
\newblock Unified vision-language pre-training for image captioning and vqa.
\newblock In \emph{AAAI}, volume~34, pages 13041--13049.

\bibitem[{Zhu et~al.(2021)Zhu, Liang, Zhu, Yu, Chang, and Liang}]{zhu2021soon}
Fengda Zhu, Xiwen Liang, Yi~Zhu, Qizhi Yu, Xiaojun Chang, and Xiaodan Liang.
  2021.
\newblock Soon: scenario oriented object navigation with graph-based
  exploration.
\newblock In \emph{CVPR}, pages 12689--12699.

\bibitem[{Zhu et~al.(2020)Zhu, Zhu, Chang, and Liang}]{zhu2020vision}
Fengda Zhu, Yi~Zhu, Xiaojun Chang, and Xiaodan Liang. 2020.
\newblock Vision-language navigation with self-supervised auxiliary reasoning
  tasks.
\newblock In \emph{CVPR}, pages 10012--10022.

\end{thebibliography}
\bibliographystyle{acl_natbib}

\clearpage
\appendix

\section{Appendix}
\label{sec:appendix}
In the Appendix, we present additional statistics and examples of our generated data. Then we discuss implementation details of prompt-based architecture.

\subsection{Dataset Details}
\begin{figure}[!t]
    \begin{center}
        \includegraphics[width=0.99\linewidth]{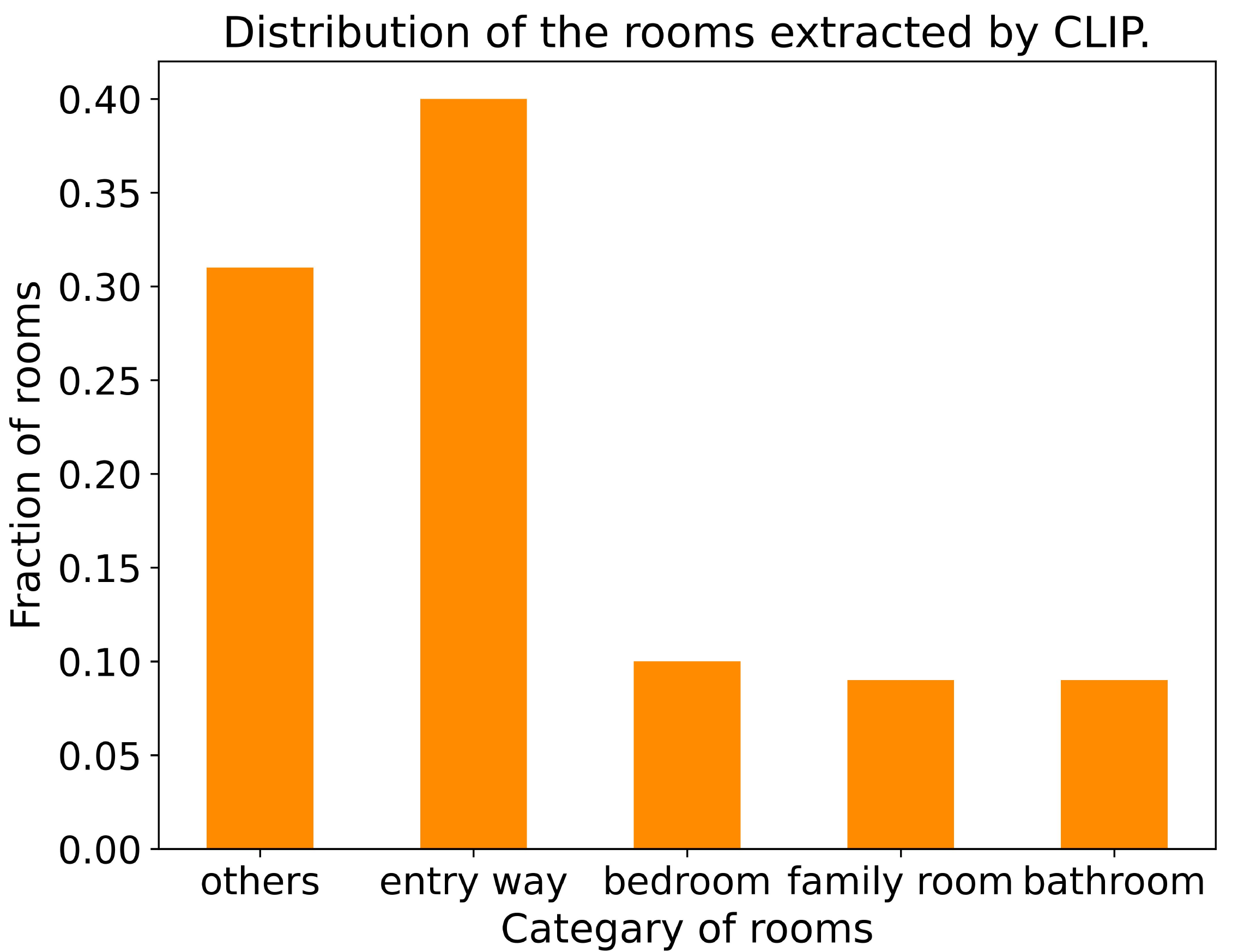} %
    \end{center}
\vspace{-2mm}
\caption{Statistical analysis of generated instructions.}
\label{fig:statistics_part2_1}
\vspace{-0.8em}
\end{figure}

\begin{figure}[!t]
    \begin{center}
        \includegraphics[width=0.99\linewidth]{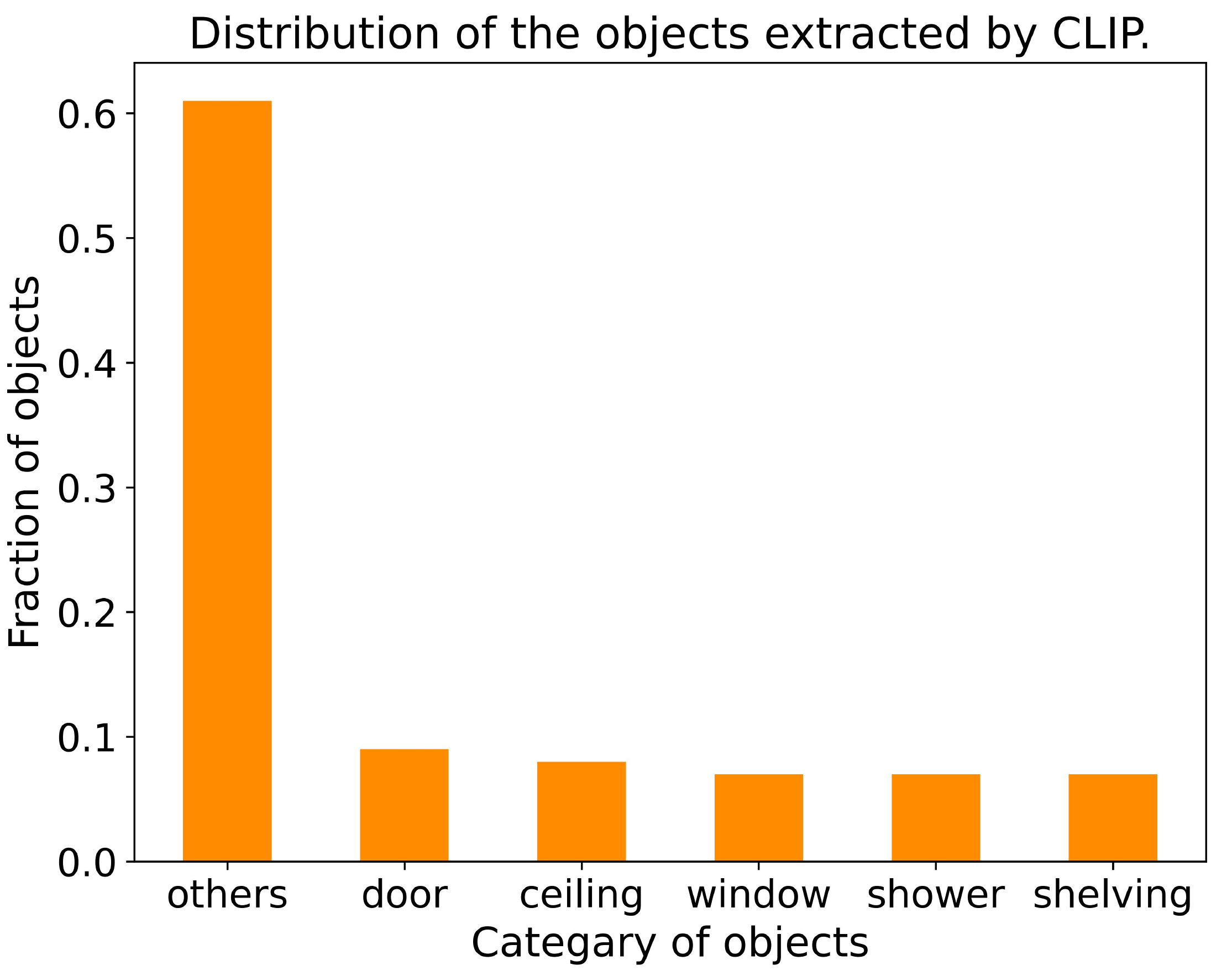} %
    \end{center}
\vspace{-2mm}
\caption{Statistical analysis of generated instructions.}
\label{fig:statistics_part2_2}
\vspace{-0.8em}
\end{figure}

\noindent\textbf{Additional Statistics}
As shown in Figure~\ref{fig:statistics_part2_1} and Figure~\ref{fig:statistics_part2_2}, we summarise rooms and objects detected by CLIP in viewpoints of sampled trajectories.
These rooms and objects appear in the indoor environment commonly, indicating the accuracy of the CLIP model.

\noindent\textbf{}

\begin{figure*}[!t]
    \begin{center}
        \includegraphics[width=0.99\linewidth]{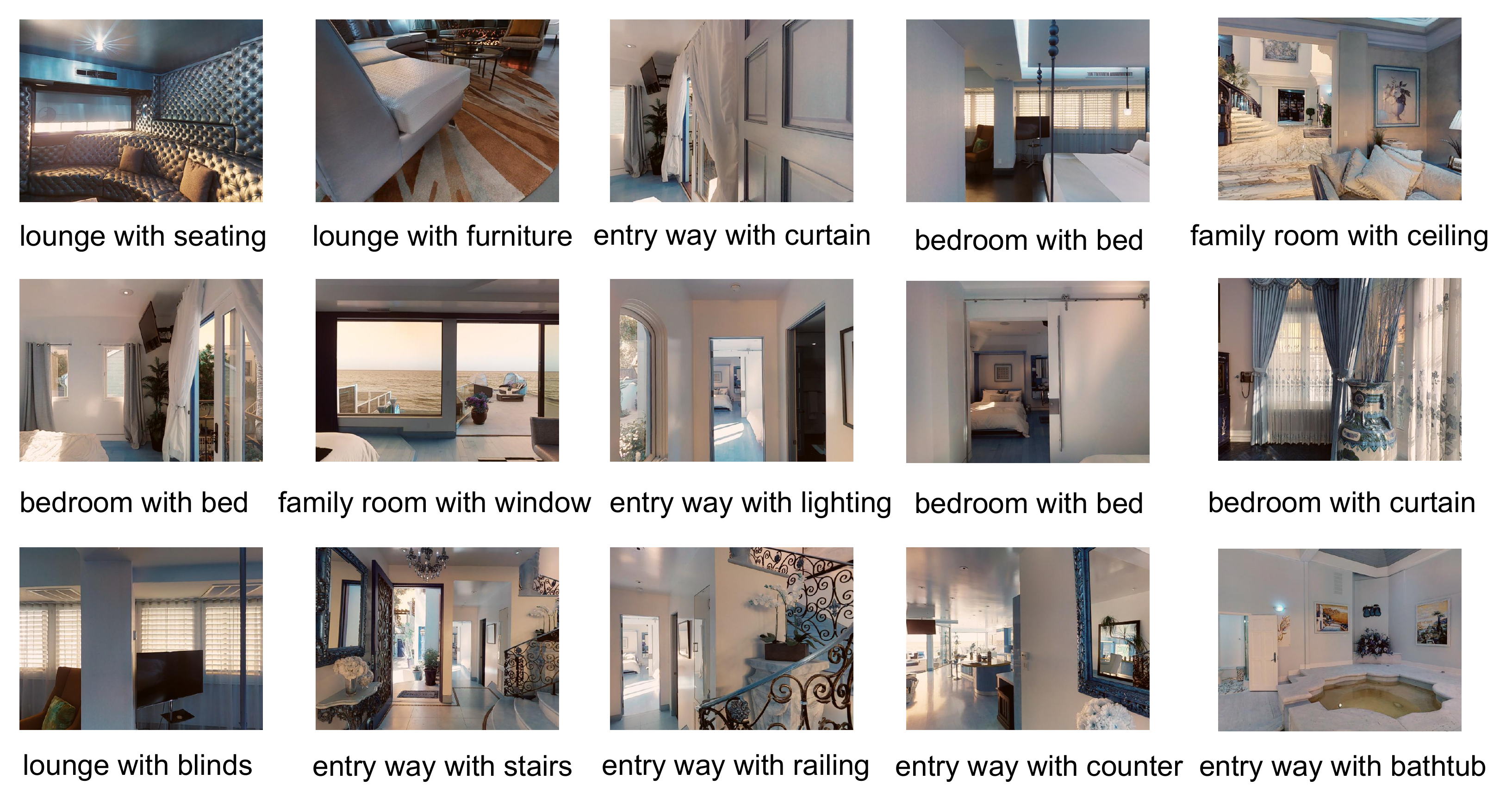}
    \end{center}
\vspace{-3mm}
\caption{Visualization of Captions.}
\label{fig:visualization of cap}
\vspace{-0.8em}
\end{figure*}

\noindent\textbf{Visualization of Captions}
We visualize generated captions for sampled viewpoints in Figure~\ref{fig:visualization of cap}. 
We infer from the figure that the CLIP can identify scenes and prominent objects accurately. 
Our generated captions contain rich visual information, 
which improves the image-text alignment ability of the model.

\begin{figure*}[!t]
    \begin{center}
        \includegraphics[width=0.95\linewidth]{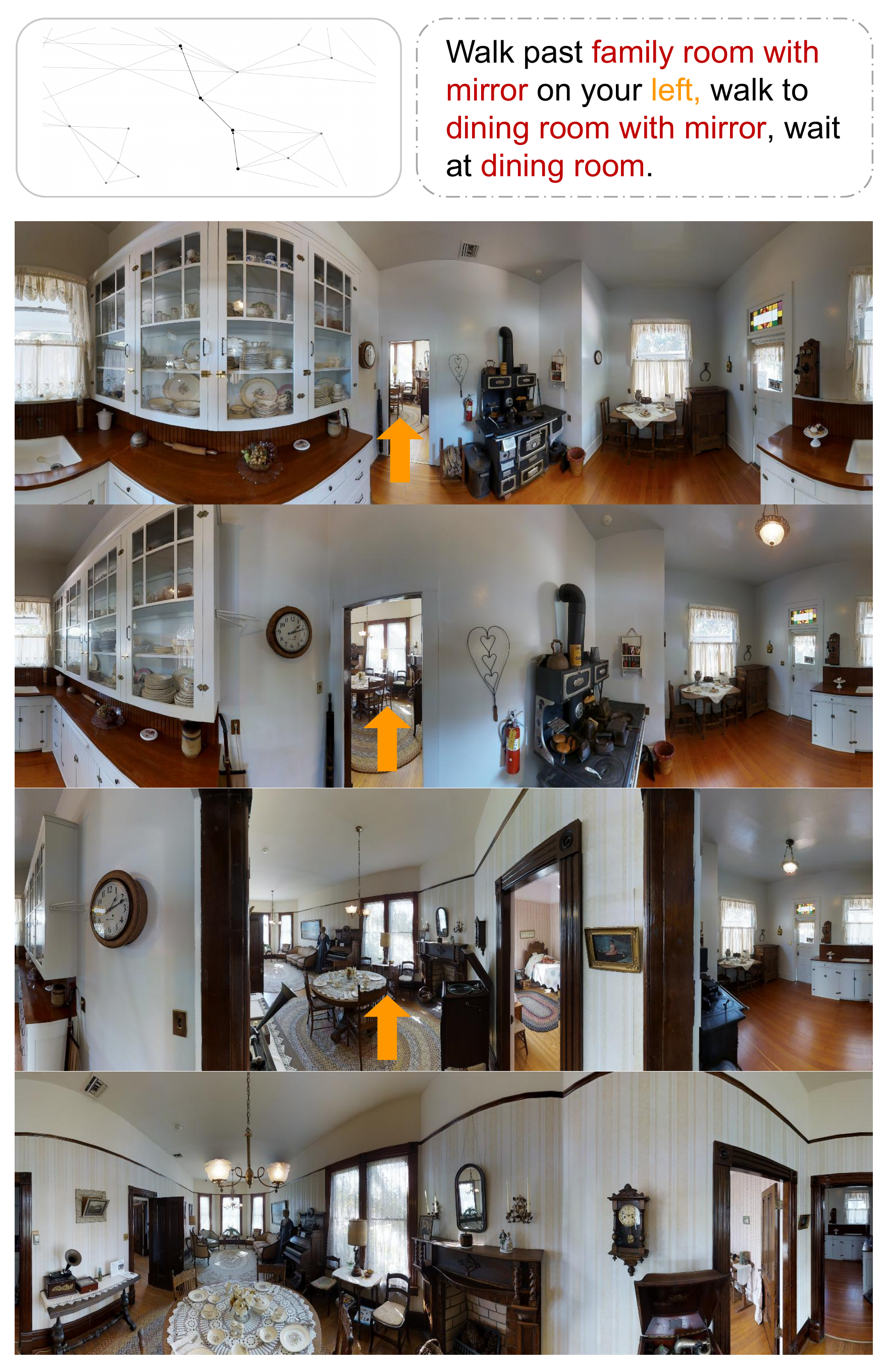}
    \end{center}
\vspace{-3mm}
\caption{Visualization of a trajectory-instruction sample generated by ProbES.}
\label{fig:vis_example1}
\vspace{-0.8em}
\end{figure*}

\begin{figure*}[!t]
    \begin{center}
        \includegraphics[width=0.95\linewidth]{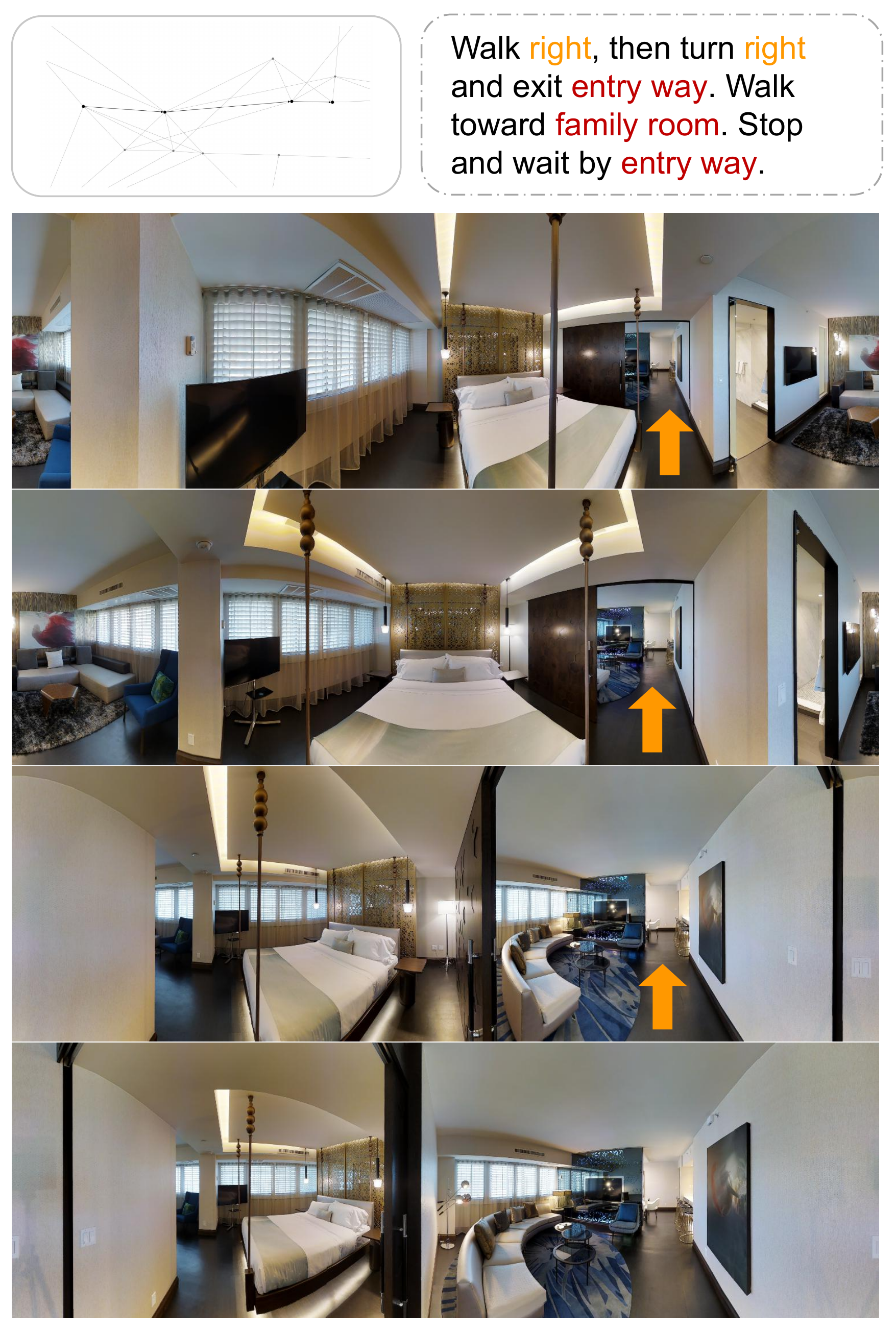}
    \end{center}
\vspace{-3mm}
\caption{Visualization of a trajectory-instruction sample generated by ProbES.}
\label{fig:vis_example2}
\vspace{-0.8em}
\end{figure*}

\noindent\textbf{Visualization of More Examples}
More examples of sampled trajectories and the corresponding generated instructions are shown in Figure~\ref{fig:vis_example1} and Figure~\ref{fig:vis_example2}, which implies that our method can generate scenario-specific instructions automatically.

\subsection{Architecture Details}
We present implementation details of our proposed prompt-based architecture for both prompt tuning in the discriminative setting and finetuning in the generative setting, respectively.

\subsubsection{Prompt-based Pretraining}
\begin{figure*}[!t]
    \begin{center}
        \includegraphics[width=0.85\linewidth]{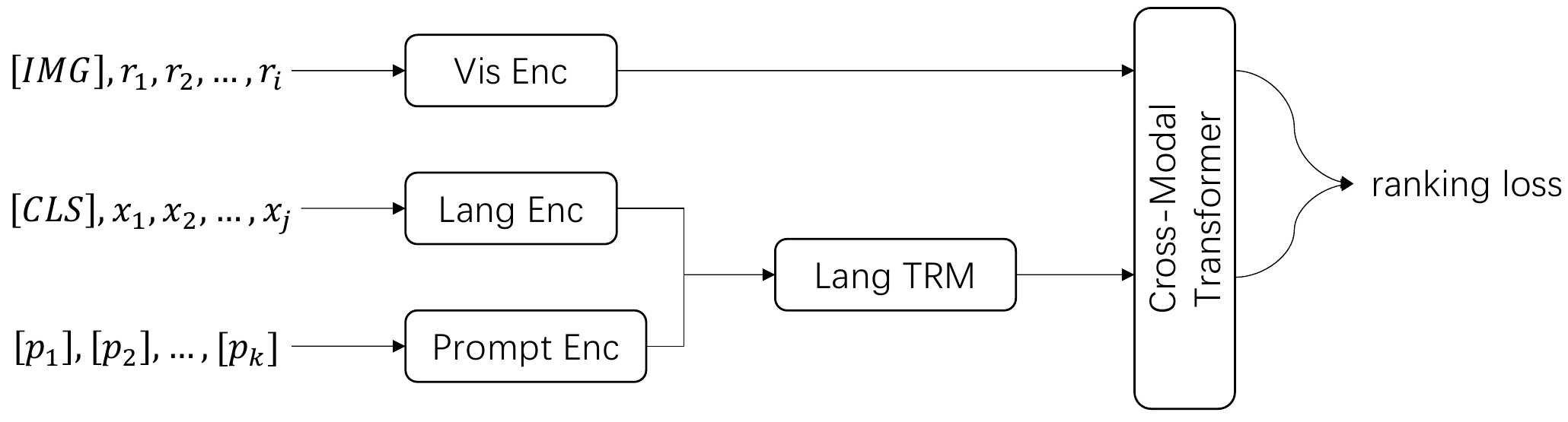}
    \end{center}
\vspace{-3mm}
\caption{Prompt tuning in discriminative setting.}
\label{fig:arc_pretrain}
\vspace{-0.5em}
\end{figure*}

As shown in Figure~\ref{fig:arc_pretrain}, the model is composed of a prompt encoder and a ViLBERT-like architecture. The prompt encoder consists of a bidirectional long-short term memory network (LSTM) and a ReLU activated two-layer multilayer perceptron (MLP). The output of the prompt encoder is prepended to the textual embedding. The ViLBERT-like architecture is similar to that of VLN-BERT. We choose ranking loss for the prompt tuning.

\subsubsection{Finetuning in Generative Setting}
\begin{figure*}[!t]
    \begin{center}
        \includegraphics[width=0.85\linewidth]{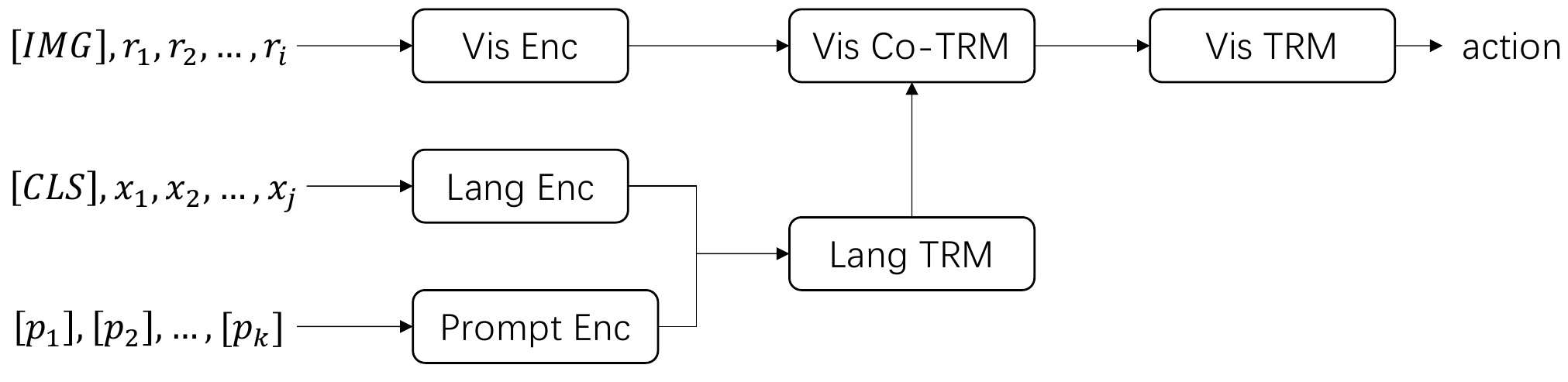}
    \end{center}
\vspace{-3mm}
\caption{Finetuning in generative setting.}
\label{fig:arc_generative}
\vspace{-0.8em}
\end{figure*}

As shown in Figure~\ref{fig:arc_generative}, the generative setting is similar to Recurrent VLN-BERT. Unlike Recurrent VLN-BERT, we introduce the prompt encoder, whose architecture is the same as the pretraining phase. During finetuning, the whole model is unfixed to achieve better results.

\end{document}